\documentclass[runningheads]{llncs}
\usepackage{graphicx}
\usepackage{todonotes}
\usepackage{multirow}

\usepackage{array}
\usepackage{tabu}
\newcolumntype{M}[1]{>{\centering\arraybackslash}m{#1}}
\newcolumntype{L}[1]{>{\arraybackslash}m{#1}}

\begin{document}
\title{Overview of BioASQ 2020: The eighth BioASQ challenge on Large-Scale Biomedical Semantic Indexing and Question Answering
}
\titlerunning{Overview of BioASQ 2020}
%

\author{
Anastasios Nentidis\inst{1,2} \and
Anastasia Krithara\inst{1} \and 
Konstantinos Bougiatiotis\inst{1,3} \and
Martin Krallinger\inst{4} \and
Carlos Rodriguez-Penagos\inst{4} \and
Marta Villegas\inst{4}\and
Georgios Paliouras\inst{1}
}
\authorrunning{A. Nentidis et al.}
%
\institute{
National Center for Scientific Research ``Demokritos'', Athens, Greece\\
\email{\{tasosnent, akrithara, bogas.ko, paliourg\}@iit.demokritos.gr}\\
\and
Aristotle University of Thessaloniki, Thessaloniki, Greece\\ \and
National and Kapodistrian University of Athens, Athens, Greece\\ \and
Barcelona Supercomputing Center, Barcelona, Spain\\
\email{\{martin.krallinger, carlos.rodriguez1, marta.villegas\}@bsc.es}
}
\maketitle              
\begin{abstract}
In this paper, we present an overview of the eighth edition of the BioASQ challenge, which ran as a lab in the Conference and Labs of the Evaluation Forum (CLEF) 2020.
BioASQ is a series of challenges aiming at the promotion of systems and methodologies for large-scale biomedical semantic indexing and question answering. 
To this end, shared tasks are organized yearly since 2012, where different teams develop systems that compete on the same demanding benchmark datasets that represent the real information needs of experts in the biomedical domain. 
This year, the challenge has been extended with the introduction of a new task on medical semantic indexing in Spanish.
In total, 34 teams with more than 100 systems participated in the three tasks of the challenge. 
As in previous years, the results of the evaluation reveal that the top-performing systems managed to outperform the strong baselines, which suggests that state-of-the-art systems keep pushing the frontier of research through continuous improvements. \\

\end{abstract}
\section{Introduction}
This paper aims at presenting the shared tasks and the datasets of the eighth BioASQ challenge in 2020, as well as at providing an overview of the participating systems and their performance.
Towards this direction, in section~\ref{sec:tasks} we provide an overview of the shared tasks, that took place from February to May 2020, and the corresponding datasets developed for the challenge. 
In section~\ref{sec:participants}, we present a brief overview of the systems developed by the participating teams for the different tasks. 
Detailed descriptions for some of the systems are available in the proceedings of the lab. 
In section~\ref{sec:results}, we focus on evaluating the performance of the systems for each task and sub-task, using state-of-the-art evaluation measures or manual assessment.
Finally, in section~\ref{sec:conclusion}, we sum up this version  of the BioASQ challenge.

\section{Overview of the Tasks}
\label{sec:tasks}
This year, the eighth version of the BioASQ challenge comprised three tasks: (1) a large-scale
biomedical semantic indexing task (task 8a), (2) a biomedical question answering task (task 8b), both considering documents in English, and (3) a new task on medical semantic indexing in Spanish (task MESINESP). In this section we provide a brief description of the two established tasks with focus on differences from previous versions of the challenge~\cite{Nentidis2019}. A detailed overview of these tasks and the general structure of BioASQ are available in \cite{Tsatsaronis2015}. In addition, we describe the new MESINESP task on semantic indexing of medical content written in Spanish (medical literature abstracts, clinical trial summaries and health-related project descriptions), which was introduced this year~\cite{Krallinger2020}, providing statistics about the dataset developed for it.

\subsection{Large-scale semantic indexing - Task 8a}

\begin{table}[!htb]
        \centering
    \begin{tabular}{M{0.1\linewidth}M{0.15\linewidth}M{0.3\linewidth}M{0.3\linewidth}}\hline
        \textbf{Batch} & \textbf{Articles} & \textbf{Annotated Articles} & \textbf{Labels per Article}  \\ \hline
        \multirow{5}{*}{1}        & 6510       & 6487                          & 12.49                             \\
                              & 7126       & 7074                          & 12.27                             \\
                              & 10891      & 10789                         & 12.55                             \\
                              & 6225       & 6182                          & 12.28                             \\
                              & 6953       & 6887                          & 12.75                             \\ \hline

    Total                  & 37705      & 37419                         & 0.99                              \\ \hline
    \multirow{5}{*}{2}        & 6815       & 6787                          & 12.49                             \\
                              & 6485       & 6414                          & 12.52                             \\
                              & 7014       & 6975                          & 11.92                             \\
                              & 6726       & 6647                          & 12.90                             \\
                              & 6379       & 6246                          & 12.45                             \\  \hline

    Total                  & 33419      & 33069                         & 0.99                              \\ \hline
    \multirow{5}{*}{3}        & 6842       & 6601                          & 12.70                             \\
                              & 7212       & 6456                          & 12.37                             \\
                              & 5430       & 4764                          & 12.59                             \\
                              & 6022       & 4858                          & 12.33                             \\
                              & 5936       & 3999                          & 12.21                             \\ \hline
    Total                  & 31442      & 26678                         & 0.85 \\  \hline
    
    \end{tabular}
        \caption{Statistics on test datasets for Task 8a.}\label{tab:a_data}
\end{table}

In Task 8a the aim is to classify articles from the PubMed/MedLine\footnote{https://pubmed.ncbi.nlm.nih.gov/} digital library into concepts of the MeSH hierarchy. In particular, new PubMed articles that are not yet annotated by the indexers in NLM are gathered to form the test sets for the evaluation of the participating systems. 
Some basic details about each test set and batch are provided in Table \ref{tab:a_data}. 
As done in previous versions of the task, the task is divided into three independent batches of 5 weekly test sets each, providing an on-line and large-scale scenario, and the test sets consist of new articles without any restriction on the journal published. 
The performance of the participating systems is calculated using standard flat information retrieval measures, as well as, hierarchical ones, when the annotations from the NLM indexers become available. 
As usual, participants have 21 hours to provide their answers for each test set. 
However, as it has been observed that new MeSH annotations are released in PubMed earlier that in previous years, we shifted the submission period accordingly to avoid having some annotations available from NLM while the task is still running.
For training, a dataset of 14,913,939 articles with 12.68 labels per article, on average, was provided to the participants.

\subsection{Biomedical semantic QA - Task 8b}

Task 8b aims at providing a realistic large-scale question answering challenge offering to the participating teams the opportunity to develop systems for all the stages of question answering in the biomedical domain. Four types of questions are considered in the task: “yes/no”, “factoid”, “list” and “summary” questions \cite{balikas13}.
A training dataset of 3,243 questions annotated with golden relevant elements and answers is provided for the participants to develop their systems.
Table \ref{tab:b_data} presents some statistics about the training dataset as well as the five test sets.

\begin{table}[!htb]
        \centering
        \begin{tabular}{M{0.1\linewidth}M{0.08\linewidth}M{0.08\linewidth}M{0.09\linewidth}M{0.1\linewidth}M{0.16\linewidth}M{0.16\linewidth}M{0.16\linewidth}}\hline
        \textbf{Batch} 	& \textbf{Size} 	&	\textbf{Yes/No}	&\textbf{List}	&\textbf{Factoid}	&\textbf{Summary}& \textbf{Documents} 	& \textbf{Snippets}  	\\ \hline
        Train       		&		3,243		&	881		&644	&941		&777	&		10.15			&	12.92			 	\\
        Test 1		&		100			&	25		&20		&32			&23		&		3.45			&	4.51			 	\\
        Test 2		&		100			&	36		&14		&25			&25		&		3.86			&	5.05			 	\\
        Test 3		&		100			&	31		&12		&28			&29		&		3.35			&	4.71			 	\\ 
        Test 4		&		100			&	26		&17		&34			&23		&		3.23			&	4.38			 	\\
        Test 5		&		100			&	34		&12		&32			&22		&		2.57			&	3.20			 	\\ \hline                    
        \textbf{Total}		&		3,743		&	1033	&719	&1092		&899	&		9.23			&	11.78			 	\\ \hline 
        \end{tabular}
        \caption{Statistics on the training and test datasets of Task 8b. The numbers for the documents and snippets refer to averages per question.}\label{tab:b_data}
\end{table}

As in previous versions of the challenge, the task is structured into two phases that focus on the retrieval of the required information (phase A) and answering the question (phase B).  
In addition, the task is split into five independent bi-weekly batches and the two phases for each batch run during two consecutive days. In each phase, the participants receive the corresponding test set and have 24 hours to submit the answers of their systems.
In particular, in phase A, a test set of 100 questions written in English is released and the participants are expected to identify and submit relevant elements from designated resources, including PubMed/MedLine articles, snippets extracted from these articles, concepts and RDF triples. 
In phase B, the manually selected relevant articles and snippets for these 100 questions are also released and the participating systems are asked to respond with \textit{exact answers}, that is entity names or short phrases, and \textit{ideal answers}, that is natural language summaries of the requested information.

\subsection{Medical semantic indexing in Spanish - MESINESP8}

There is a pressing need to improve the access to information comprised in health and biomedicine related documents, not only by professional medical users buy also by researches, public healthcare decision makers, pharma industry and particularly by patients. Currently, most of the Biomedical NLP and IR research is being done on content in English, despite the fact that a large volume of medical documents is published in other languages including Spanish. Key resources like PubMed focus primarily on data in English, but it provides outlinks also to articles originally published in Spanish. MESINESP attempts to promote the development of systems for automatic indexing with structured medical vocabularies (DeCS terms) of healthcare content in Spanish: IBECS\footnote{\footnotesize IBECS includes bibliographic references from scientific articles in health sciences published in Spanish journals. \url{http://ibecs.isciii.es}}, LILACS\footnote{\footnotesize LILACS is the most important and comprehensive index of scientific and technical literature of Latin America and the Caribbean. It includes 26 countries, 882 journals and 878,285 records, 464,451 of which are full texts \url{https://lilacs.bvsalud.org}}, REEC 
\footnote{\footnotesize Registro Español de Estudios Clínicos, a database containing summaries of clinical trials \url{https://reec.aemps.es/reec/public/web.html}} and  FIS-ISCIII
\footnote{\footnotesize public healthcare project proposal summaries (Proyectos de Investigación en Salud, diseñado por el Instituto de Salud Carlos III, ISCIII) \url{https://portalfis.isciii.es/es/Paginas/inicio.aspx}}. The main aim of MESINESP is to promote the development of semantic indexing tools of practical relevance of non-English content, determining the current-state-of-the art, identifying challenges and comparing the strategies and results to those published for English data. This task was organized within the framework of the Spanish Government's Plan for Promoting Language Technologies (Plan TL), that aims to promote the development of natural language processing, machine translation and conversational systems in Spanish and co-official languages.

A training dataset with 369,368 articles manually annotated with DeCS codes (\emph{Descriptores en Ciencias de la Salud}, derived and extended from MeSH terms)\footnote{\footnotesize 29,716 come directly from MeSH and 4,402 are exclusive to DeCS } was released. 1,500 articles were manually annotated and verified at least by two human experts (from a pool of 7 annotators), and from them a development and gold standard for evaluation  were generated. A further background dataset was produced from diverse sources, including machine-translated text.
Consistently, the different collections averaged, per document, around 10 sentences, 13 DeCS codes, and 300 words, of which between 130 and 140 were unique ones. 

\begin{figure*}[!htb]
\centerline{\includegraphics[width=1\textwidth]{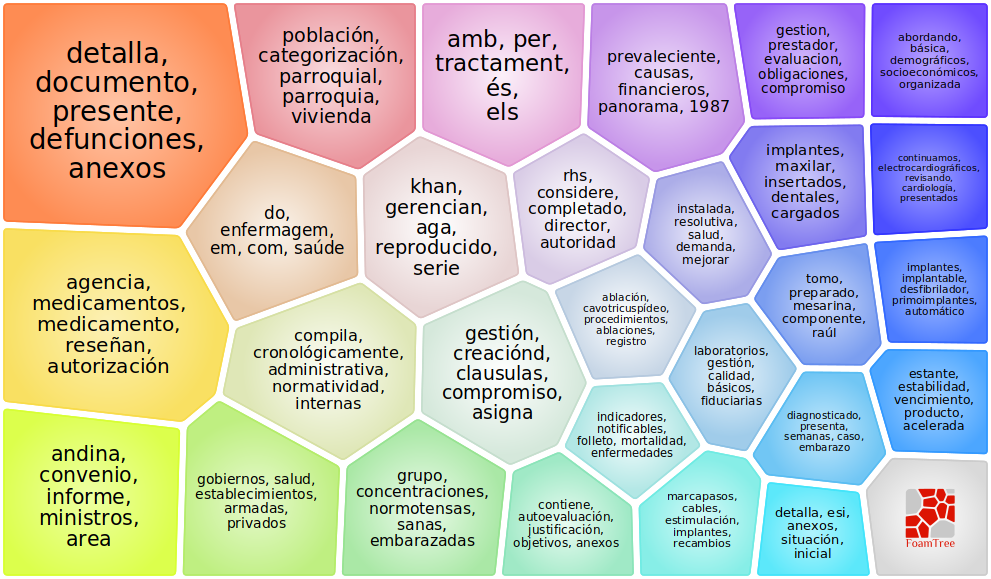}}
\caption{Content visualization of MESINESP training dataset using clustering techniques. Among subjects shown: clinical cases, non-Spanish languages, medication and device reviews, health care management etc. This reflects DeCS extension from MeSH terms to other subjects, such as Public Health issues. }\label{fig:02}
\end{figure*}

In order to explore the diversity of content from this dataset, we generated clusters of semantically similar records from the training dataset's titles by, first, creating a Doc2Vec model with the gensim library,\footnote{\footnotesize \url{https://radimrehurek.com/gensim/}} and then using that similarity matrix to feed an unsupervised DBScan algorithm from the sklearn python package,\footnote{\footnotesize \url{https://scikit-learn.org/}} that basically creates clusters from high density samples. The resulting 27 clusters were visualized with the libraries from the Carrot Workbench project. \footnote{\footnotesize \url{https://project.carrot2.org/}}
 (Figure {\ref{fig:02}}).

\section{Overview of participation}
\label{sec:participants}
\subsection{Task 8a}
This year, 7 teams participated in the eighth edition of task a, submitting predictions from 16 different systems in total. Here, we provide a brief overview of those systems for which a description was available,
stressing their key characteristics. A summing-up of the participating systems and corresponding approaches is presented in Table~\ref{tab:a_sys}.

\begin{table}[!htb]
        \centering
        \begin{tabular}{M{0.3\linewidth}M{0.6\linewidth}}\hline
        \textbf{System} & \textbf{Approach} \\ \hline
        X-BERT BioASQ & X-BERT, Transformers ELMo, MER \\\hline
        NLM CNN & SentencePiece, CNN, embeddings, ensembles\\\hline
        dmiip\_fdu & d2v, tf-idf, SVM, KNN, LTR,  DeepMeSH, AttentionXML, BERT, PLT \\\hline
        Iria & Luchene Index, k-NN, stem bigrams, ensembles, UIMA ConceptMapper\\\hline

        \end{tabular}
         \caption{Systems and approaches for Task 8a. Systems for which no description was available at the time of writing are omitted. }\label{tab:a_sys}
\end{table}

This year, the LASIGE team from the University of Lisboa, in its ``X-BERT BioASQ'' system propose a  novel approach for biomedical semantic indexing combining a solution based on Extreme Multi-Label Classification (XMLC) with a Named-Entity-Recognition (NER) tool. 
In particular, their system is based on X-BERT~\cite{chang2019x}, an approach to scale BERT~\cite{Devlin2018} to XMLC, combined with the use of the MER~\cite{Couto2018} tool to recognize MeSH terms in the abstracts of the articles.
The system is structured into three steps. The first step is the semantic indexing of the labels into clusters using ELMo~\cite{Peters2018}; then a second step matches the indices using a Transformer architecture; and finally, the third step focuses on ranking the labels retrieved from the previous indices.

Other teams, improved upon existing systems already participating in previous versions of the task. 
Namely, the National Library of Medicine (NLM) team, in its ``\textit{NLM CNN}'' system enhance the previous version of their ``\textit{ceb}'' systems \cite{Rae2019}, based on an end-to-end Deep Learning (DL) architecture with Convolutional Neural Networks (CNN), with SentencePiece tokenization~\cite{Kudo2018}.
The Fudan University team also builds upon their previous ``\textit{AttentionXML}''~\cite{You2018} and ``\textit{DeepMeSH}''~\cite{peng2016} systems as well their new ``\textit{BERTMeSH}'' system, which are based on document to vector (d2v) and tf-idf feature embeddings, learning to rank (LTR) and DL-based extreme multi-label text classification, Attention Mechanisms and Probabilistic Label Trees (PLT)~\cite{Jain2016}. 
Finally, this years versions of the ``\textit{Iria}'' systems~\cite{Ribadas2015} are also based on the same techniques used by the systems in previous versions of the challenge which are summarized in Table~\ref{tab:a_sys}. 

Similarly to the previous versions of the challenge, two systems developed by NLM to facilitate the annotation of articles by indexers in MedLine/PubMed, where available as baselines for the semantic indexing task. MTI \cite{morkBioasq2014} as enhanced in \cite{zavorin2016} and an extension based on features suggested by the winners of the first version of the task \cite{tsoumakasBioasq}.

\subsection{Task 8b}

This version of Task b was tackled by 94 different systems in total, developed by 23 teams. In particular, 8 teams participated in the first phase, on the retrieval of relevant material required for answering the questions, submitting results from 30 systems. 
In the second phase, on providing the exact and ideal answers for the questions, participated 18 teams with 72 distinct systems. 
Three of the teams participated in both phases. 
An overview of the technologies employed by the teams is provided in Table \ref{tab:b_sys} for the systems for which a description were available. Detailed descriptions for some of the systems are available at the proceedings of the workshop.

\begin{table}[!htb]
        \centering
        \begin{tabular}{M{0.2\linewidth}M{0.1\linewidth}M{0.55\linewidth}}\hline
        \textbf{Systems} & \textbf{Phase}& \textbf{Approach} \\ \hline
        pa & A, B & 
            BM25, BERT, 
            Word2Vec, 
            SQuAD, PubMedQA, 
            BioMed-RoBERTa 
        \\\hline 
        bio-answerfinder & A, B &  
            Bio-AnswerFinder, LSTM, ElasticSearch, BERT, 
            Electra, BioBERT, SQuAD, wRWMD
        \\\hline 
        Google & A & BM25, BioBERT, Synthetic  Query  Generation, BERT, reranking  \\\hline
        bioinfo & A & BM25, ElasticSearch, distant learning, DeepRank  \\\hline
        KU-DMIS & B & 
            BioBERT, NLI, MultiNLI, SQuAD, 
            BART, beam search, BERN, language\_check
        \\\hline
        NCU-IISR & B & BioBERT, logistic regression, LTR \\\hline 

        UoT & B & BioBERT, multi-task learning, BC2GM \\\hline
        BioNLPer & B & BioBERT, multi-task learning, NLTK, ScispaCy \\\hline
        LabZhu & B & BERT, BoiBERT, XLNet, SpanBERT, transfer learning, SQuAD, ensembling\\\hline    
        MQ & B & Word2Vec, BERT, LSTM, Reinforcement Learning (PPO) \\\hline 
        DAIICT & B & textrank, lexrank, UMLS \\\hline
        sbert & B & Sentence-BERT, BioBERT, SNLI, MutiNLI, multi-task learning, MQU \\\hline
        \hline
        \end{tabular}
        \caption{Systems and approaches for Task8b. Systems for which no information was available at the time of writing are omitted.}\label{tab:b_sys}
\end{table}

The ``\textit{ITMO}'' team participated in both phases of the task experimenting in its ``pa'' systems with  differing solutions across the batches. In general, for document retrieval the systems follow an approach with two stages. First, they identify initial candidate articles based on BM25, and then they re-rank them using variations of BERT~\cite{Devlin2018}, fine-tuned for the binary classification task with the BioASQ dataset and pseudo-negative documents. They extract snippets from the top documents and rerank them using biomedical Word2Vec based on cosine similarity with the question. To extract exact answers they use BERT fine-tuned on SQUAD~\cite{rajpurkar2016squad} and BioASQ datasets and employ a post-processing to split the answer for list questions and additional fine-tuning on PubMedQA~\cite{jin2019pubmedqa} for yes/no questions. 
Finally, for ideal answers they generate some candidates from the snippets and their sentences and rerank them using the model used for phase A. In the last batch, they also experiment with generative summarization, developing a model based on BioMed-RoBERTa~\cite{gururangan2020don} to improve the readability and consistency of the produced ideal answers. 

Another team participating in both phases of the task is the ``\textit{UCSD}'' team with its ``bio-answerfinder'' system. In particular, for phase A they rely on previously developed Bio-AnswerFinder system~\cite{ozyurt2020bio}, which is also used as a first step in phase B, for re-ranking the sentences of the snippets provided in the test set.
For identifying the exact answers for factoid and list questions they experimented on fine-tuning Electra~\cite{clark2020electra} and BioBERT~\cite{lee2019biobert} on SQuAD and BioASQ datasets combined.
The  answer candidates are then scored considering classification probability, the top ranking of corresponding snippets and number of occurrences. Finally a normalization and filtering step is performed and, for list questions, and enrichment step based on coordinated phrase detection.
For yes/no questions they fine-tune BioBERT on the BioASQ dataset and use majority voting.
For summary questions, they employ hierarchical clustering, based on weighted relaxed word mover's distance (wRWMD) similarity~\cite{ozyurt2020bio} to group the top sentences, and select the sentence ranked highest by Bio-AnswerFinder to be concatenated to form the summary. 

In phase A, the ``\textit{Google}'' team participated with four distinct systems based on different approaches. In particular, they used a BM25 retrieval model, a neural retrieval model, initialized with BioBERT and trained on a large set of questions developed through Synthetic Query Generation (QGen), and a hybrid retrieval model~\footnote{https://ai.googleblog.com/2020/05/an-nlu-powered-tool-to-explore-covid-19.html} based on a linear blend of BM25 and the neural model~\cite{ma2020zero}. In addition, they also used a reranking model, rescoring the results of the hybrid model with a cross-attention BERT rescorer~\cite{pappas2019}.
The team from the University of Aveiro, also participated in phase A with its ``bioinfo'' systems, which consists of a fine-tuned BM25 retrieval model based on ElasticSearch~\cite{gormley2015elasticsearch}, followed by a neural reranking step. For the latter, they use an interaction-based model inspired on the DeepRank~\cite{pang2017deeprank} architecture building upon previous versions of their system~\cite{almeida2020calling}. The focus of the improvements was on the sentence splitting strategy, on extracting of multiple relevance signals, and the independent contribution of each sentence for the final score.

In phase B, this year the ``\textit{KU-DMIS}'' team participated on both exact and ideal answers.
For exact answers, they build upon their previous BioBERT-based systems~\cite{Yoon2019} and try to adapt the sequential transfer learning of Natural Language Inference (NLI) to biomedical question answering. In particular, they investigate whether learning knowledge of entailment between two sentence pairs can improve exact answer generation, enhancing their BioBERT-based models with alternative fine-tuning configurations based on the MultiNLI dataset~\cite{williams2017broad}.
For ideal answer generation, they develop a deep neural abstractive summarization model based on BART~\cite{lewis2019bart} and beam search, with particular focus on pre-processing and post-processing steps. In particular, alternative systems were developed either considering the answers predicted by the exact answer prediction system in their input or not. 
In the post-processing step, the generated candidate ideal answers for each question where scored using the predicted exact answers and some grammar scores provided by the language\_check tool\footnote{https://pypi.org/project/language-check/}.
For factoid and list questions in particular, the BERN~\cite{kim2019neural} tool was also employed to recognize named entities in the candidate ideal answers for the scoring step.  

The ``\textit{NCU-IISR}'' team also participated in both parts of phase B, constructing two BioBERT-based models for extracting the exact answer and ranking the ideal answers respectively. The first model is fine-tuned on the BioASQ dataset formulated as a SQuAD-type QA task that extracts the answer span. For the second model, they regard the sentences of the provided snippets as candidate ideal answers and build a ranking model with two parts. First, a BioBERT-based model takes as input the question and one of the snippet sentences and provides their representation. Then, a logistic regressor, trained on predicting the similarity between a question and each snippet sentence, takes this representation and outputs a score, which is used for selecting the final ideal answer. 

The ``\textit{UoT}'' team participated with three different DL approaches for generating exact answers. In their first approach, they fine-tune separately two distinct BioBERT-based models extended with an additional neural layer depending on the question type, one for yes/no and one for factoid and list questions together. In their second system, they use a joint-learning setting, where the same BioBERT layer is connected with both the additional layers and jointly trained for all types of questions.   
Finally, in their third system they propose a multi-task model to learn recognizing biomedical entities and answers to questions simultaneously, aiming at transferring knowledge from the biomedical entity recognition task to question answering. In particular, they 
extend their joint BioBERT-based model with simultaneous training on the BC2GM dataset~\cite{smith2008overview} for recognizing gene and protein entities.

The ``\textit{BioNLPer}'' team also participated in the exact answers part of phase B, focusing on factoids. They proposed 5 BioBERT-based systems, using external feature enhancement and auxiliary task methodologies. 
In particular, in their ``factoid qa model'' and ``Parameters retrained'' systems they consider the prediction of answer boundaries (start and end positions) as the main task and the whole answer content prediction as an auxiliary task.
In their ``Features Fusion'' system they leveraged external features including NER and part-of-speach (POS) extracted by NLTK~\cite{loper2002nltk} and ScispaCy~\cite{neumann2019scispacy} tools as additional textual information and fused them with the pre-trained language model representations, to improve answer boundary prediction. 
Then, in their ``BioFusion'' system they combine the two  methodologies together.
Finally, their ``BioLabel'' system employed the general and biomedical domain corpus classification as the auxiliary task to help answer boundary prediction.

The ``LabZhu'' systems also participated in phase B, with focus on exact answers for the factoid and list questions. They treat answer generation as an extractive machine comprehension task and explore several different pretrained language models, including BERT, BioBERT, XLNet~\cite{DBLP:journals/corr/abs-1906-08237} and SpanBERT~\cite{joshi2020spanbert}. They also follow a transfer learning approach, training the models on the SQuAD dataset, and then fine-tuning them on the BioASQ datasets. Finally, they also rely on  voting to integrate the results of multiple models.

The ``\textit{MQ}'' team, as in past years, focused on ideal answers, approaching the task as query-based summarisation. In some of their systems the retrain their previous classification and regression approaches~\cite{molla2019classification} in the new training dataset. In addition, they also employ reinforcement learning with Proximal Policy Optimization (PPO)~\cite{schulman2017proximal} and two variants to represent the input features, namely Word2Vec-based and BERT-based embeddings.
The ``\textit{DAIICT}'' team also participated in ideal answer generation, using the standard extractive summarization techniques textrank~\cite{mihalcea2004textrank} and lexrank~\cite{erkan2004lexrank} as well as sentence selection techniques based on their similarity with the query. They also modified these techniques investigating the effect of query expansion based on UMLS~\cite{bodenreider2004unified} for sentence selection and summarization.

Finally, the ``\textit{sbert}'' team, also focused on ideal answers. They experimented with different embedding models and multi-task learning in their systems, using parts from previous ``\textit{MQU}'' systems for the pre-processing of data and the prediction step based on classification and regression~\cite{molla2019classification}.
In particular, they used a Universal Sentence Embedding Model~\cite{conneau2017supervised}  (BioBERT-NLI~\footnote{https://huggingface.co/gsarti/biobert-nli}) based on a version of BioBERT fine-tuned on the the SNLI~\cite{bowman2015large} and the MultiNLI datasets as in Sentence-BERT~\cite{reimers2019sentence}.
The features were fed to either a single logistic regression or classification model to derive the ideal answers. Additionally, in a multi-task setting, they trained the model on both the classification and regression tasks, selecting for the final prediction one of them.

In this challenge too, the open source OAQA system proposed by \cite{yang2016learning} served as baseline for phase B exact answers. The system which achieved among the highest performances in previous versions of the challenge remains a strong baseline for the exact answer generation task. The system is developed based on the UIMA framework. ClearNLP is employed for question and snippet parsing. MetaMap, TmTool \cite{Wei2016}, C-Value and LingPipe \cite{baldwin2003lingpipe} are used for concept identification and UMLS Terminology Services (UTS) for concept retrieval. The final steps include identification of concept, document and snippet relevance based on classifier components and scoring and finally ranking techniques.

\subsection{Task MESINESP8}
\begin{table}[!htb]
        \centering
        \begin{tabular}{M{0.3\linewidth}M{0.5\linewidth}}\hline
        \textbf{System} & \textbf{Approach} \\ \hline
        Iria & bigrams, Luchene Index, k-NN, ensembles, UIMA ConceptMapper\\\hline
        Fudan University & AttentionXML with multilingual-BERT \\\hline 
        Alara (UNED) & Frequency graph matching \\\hline
        Priberam & BERT based classifier, and SVM-rank ensemble\\\hline 
        LASIGE & X-BERT, Transformers ELMo, MER \\\hline
        \end{tabular}
         \caption{Systems and approaches for Task MESINESP8. Systems for which no description was available at the time of writing are omitted. }\label{tab:mesinesp_sys}
\end{table}
For the newly introduced MESINESP8 task, 6 teams from China, India, Portugal and Spain  participated and results from 24 different systems were submitted.
The approaches were similar to the comparable English task, and included KNN and Support Vector Machine classifiers, as well as deep learning frameworks like X-BERT and multilingual-BERT, already described in  subsection 3.1. 
 A simple lookup system was provided as a baseline for the MESINESP task. This system extracts information from an annotated list. Then checks whether, in a set of text documents, the annotation are present. It basically gets the intersection between tokens in annotations and tokens in words. This simple approach obtains a MiF of 0.2695.

\section{Results}
\label{sec:results}

\subsection{Task 8a}
\begin{table*}[!htbp]
\centering
\begin{tabular}{M{0.3\linewidth}M{0.1\linewidth}M{0.1\linewidth}M{0.1\linewidth}M{0.1\linewidth}M{0.1\linewidth}M{0.1\linewidth}}\hline
\textbf{System} & \multicolumn{2}{c}{\textbf{Batch 1}} & \multicolumn{2}{c}{\textbf{Batch 2}} & \multicolumn{2}{c}{\textbf{Batch 3}} \\ \hline
& MiF & LCA-F & MiF & LCA-F & MiF & LCA-F \\ \cline{2-7}
deepmesh\_dmiip\_fdu   & \textbf{1.25} & \textbf{2.25} & 1.875         & 3.25          & 2.25         & 2.25          \\
deepmesh\_dmiip\_fdu\_ & 2.375  & 3.625         & \textbf{1.25} & \textbf{1.25} & 1.75         & 2             \\
attention\_dmiip\_fdu  & 3      & \textbf{2.25} & 3.5           & 3.125         & 3            & 3.25          \\
Default MTI            & 4.75   & 3.75   & 6       & 5.25    & 6            & 5.5           \\
MTI First Line Index   & 5.5    & 4.5    & 6.75    & 5.875   & 5.75         & 5.25          \\
dmiip\_fdu             & -      & -      & 2.375   & 1.625   & \textbf{1.5} & \textbf{1.25} \\
NLM CNN                & -      & -      & 5       & 6.75    & 5.5      & 7        \\
iria-mix               & -      & -      & -       & -       & 8.25     & 8.25     \\
iria-1                 & -      & -      & -       & -       & 9.25     & 9.25     \\
X-BERT BioASQ          & -      & -      & -       & -       & 10.75    & 10.75    \\
\hline
\end{tabular}
\caption{Average system ranks across the batches of the task 8a. A hyphenation symbol (-) is used whenever the system participated in fewer than 4 test sets in the batch. Systems participating in fewer than 4 test sets in all three batches are omitted.}\label{tab:a_res}
\end{table*}

In Task 8a, each of the three batches were independently evaluated as presented in Table~\ref{tab:a_res}.
Standard flat and hierarchical evaluation measures \cite{balikas13} were used for measuring the classification performance of the systems. In particular, the micro F-measure (MiF) and the Lowest Common Ancestor F-measure (LCA-F) were used to identify the winners for each batch \cite{kosmopoulos2015evaluation}.
As suggested by Demšar \cite{Demsar06}, the appropriate way to compare multiple classification systems over multiple datasets is based on their average rank across all the datasets. 
In this task, the system with the best performance in a test set gets rank 1.0 for this test set, the second best rank 2.0 and so on. 
In case two or more systems tie, they all receive the average rank.
Then, according to the rules of the challenge, the average rank of each system for a batch is calculated based on the four best ranks of the system in the five test sets of the batch. 
The average rank of each system, based on both the flat MiF and the hierarchical LCA-F scores, for the three batches of the task are presented in Table~\ref{tab:a_res}. 

\begin{figure*}[!htb]
\centerline{\includegraphics[width=1\textwidth]{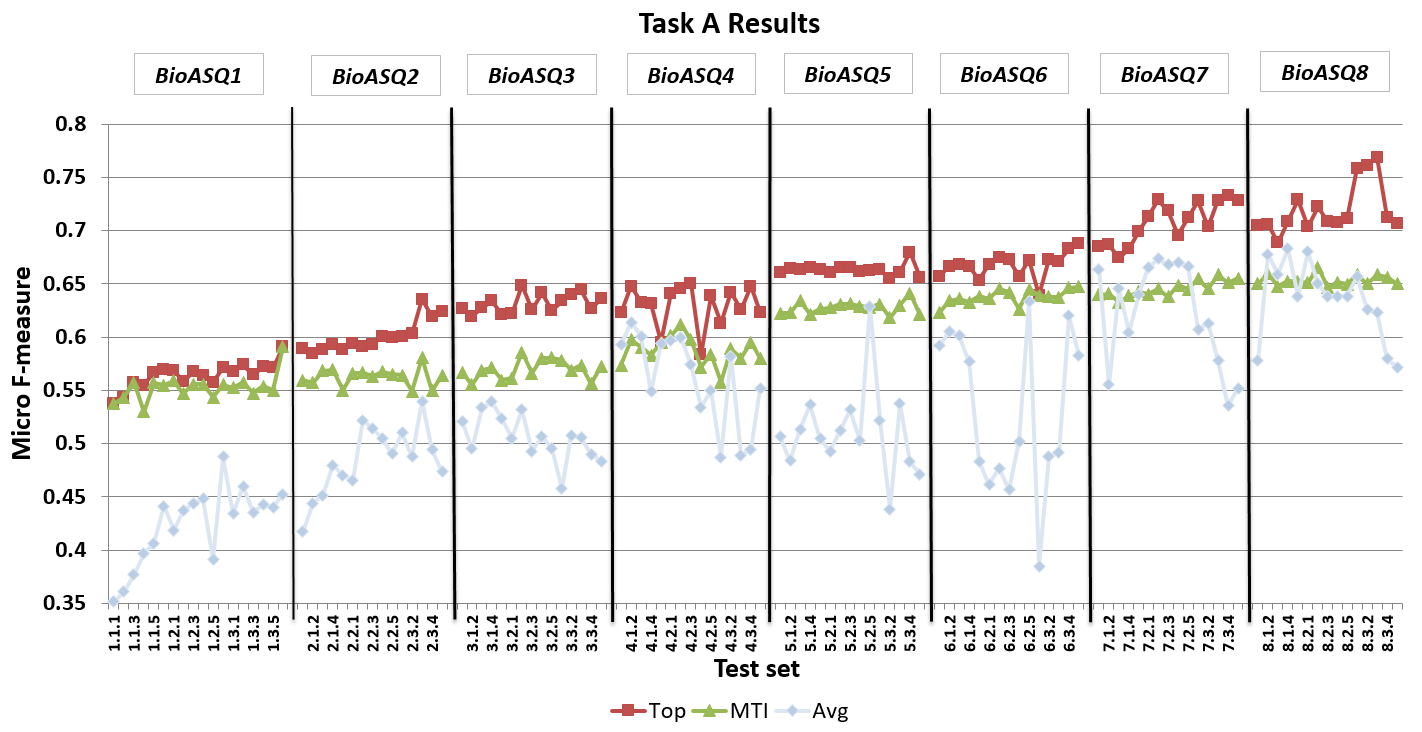}}
\caption{The micro f-measure (MiF) achieved by systems across different years of the BioASQ challenge. For each test set the MiF score is presented for the best performing system (Top) and the MTI, as well as the average micro f-measure of all the participating systems (Avg). }\label{fig:01}
\end{figure*}

\indent The results in Task 8a show that in all test batches and for both flat and hierarchical measures, the best systems outperform the strong baselines. In particular, the ``\textit{dmiip\_fdu}'' systems from the Fudan University team achieve the best performance in all three batches of the task. More detailed results can be found in the online results page\footnote{\footnotesize \url{http://participants-area.bioasq.org/results/8a/}}. 
Comparing these results with the corresponding results from previous versions of the task, suggests that both the MTI baseline and the top performing systems keep improving through the years of the challenge, as shown in Figure {\ref{fig:01}}.

\subsection{Task 8b}
\textbf{Phase A}: 
In the first phase of Task 8b, the systems are ranked according to the Mean Average Precision (MAP) measure for each of the four types of annotations, namely documents, snippets, concepts and RDF triples.
This year, the calculation of Average Precision (AP) in MAP for phase A was reconsidered as described in the official description of the evaluation measures for Task 8b\footnote{http://participants-area.bioasq.org/Tasks/b/eval\_meas\_2020/}.
In brief, since BioASQ3, the participant systems are allowed to return up to 10 relevant items (e.g. documents), and the calculation of AP was modified to reflect this change. However, the number of golden relevant items in the last years have been observed to be lower than 10 in some cases, resulting to relatively small AP values even for submissions with all the golden elements. For this reason, this year, we modified the MAP calculation to consider both the limit of 10 elements and the actual number of golden elements.  
In Tables \ref{tab:bA_res_doc} and \ref{tab:bA_res_sni}
some indicative preliminary results from batch 2 are presented. The full results are available in the online results page of Task 8b, phase A\footnote{\footnotesize \url{http://participants-area.bioasq.org/results/8b/phaseA/}}. The results presented here are preliminary, as the final results for the task 8b will be available after the manual assessment of the system responses by the BioASQ team of biomedical experts.

\begin{table*}[!htbp]
\centering
\begin{tabular}{M{0.3\linewidth}M{0.13\linewidth}M{0.13\linewidth}M{0.13\linewidth}M{0.13\linewidth}M{0.12\linewidth}}\hline
\textbf{System} & \textbf{Mean Precision} & \textbf{Mean Recall} & \textbf{Mean F-measure} & \textbf{MAP} & \textbf{GMAP}  \\ \hline
pa                   & \textbf{0.1934}	& 0.4501   & \textbf{0.2300} & \textbf{0.3304} & 0.0185\\
AUEB-System1         & 0.1688    & \textbf{0.4967} & 0.2205  & 0.3181 & 0.0165\\
bioinfo-3            & 0.1500    & 0.4880   & 0.2027   & 0.3168   & \textbf{0.0223}\\
bioinfo-1            & 0.1480    & 0.4755   & 0.1994   & 0.3149   & 0.0186   \\
bioinfo-4            & 0.1500    & 0.4787   & 0.2002   & 0.3120   & 0.0161   \\
AUEB-System2         & 0.1618    & 0.4864   & 0.2126   & 0.3103   & 0.0149   \\
bioinfo-2            & 0.1420    & 0.4648   & 0.1914   & 0.3084   & 0.0152   \\
bioinfo-0            & 0.1380    & 0.4341   & 0.1830   & 0.2910   & 0.0117   \\
AUEB-System5         & 0.1588    & 0.4549   & 0.2057   & 0.2843   & 0.0116   \\
Ir\_sys4             & 0.1190    & 0.4179   & 0.1639   & 0.2807   & 0.0056   \\
Google-AdHoc-MAGLEV  & 0.1310    & 0.4364   & 0.1770   & 0.2806   & 0.0109   \\
Ir\_sys2             & 0.1190    & 0.4179   & 0.1639   & 0.2760   & 0.0055   \\
Google-AdHoc-BM25    & 0.1324    & 0.4222   & 0.1758   & 0.2718   & 0.0088   \\
AUEB-System3         & 0.1688    & \textbf{0.4967}   & 0.2205   & 0.2702   & 0.0146   \\
Ir\_sys3             & 0.1325    & 0.3887   & 0.1730   & 0.2678   & 0.0045   \\
\hline
\end{tabular}
\caption{Results for document retrieval in batch 2 of phase A of Task 8b. 
Only the top-15 systems are presented.
}\label{tab:bA_res_doc}
\end{table*}

\begin{table*}[!htbp]
\centering
\begin{tabular}{M{0.3\linewidth}M{0.14\linewidth}M{0.12\linewidth}M{0.14\linewidth}M{0.12\linewidth}M{0.12\linewidth}}\hline
\textbf{System} & \textbf{Mean Precision} & \textbf{Mean Recall} & \textbf{Mean F-measure} & \textbf{MAP} & \textbf{GMAP}  \\ \hline

AUEB-System1         & \textbf{0.1545}                  & 0.2531          & \textbf{0.1773}             & \textbf{0.6821}       & 0.0015        \\
AUEB-System2         & 0.1386                  & 0.2260          & 0.1609             & 0.6549       & 0.0011        \\
pa                   & 0.1348                  & \textbf{0.2578}          & 0.1627             & 0.3374       & \textbf{0.0047}        \\
bioinfo-4            & 0.1308                  & 0.2009          & 0.1413             & 0.2767       & 0.0016        \\
bioinfo-1            & 0.1373                  & 0.2103          & 0.1461             & 0.2721       & 0.0016        \\
bioinfo-2            & 0.1299                  & 0.2018          & 0.1408             & 0.2637       & 0.0011        \\
bioinfo-3            & 0.1321                  & 0.2004          & 0.1404             & 0.2607       & 0.0014        \\
MindLab QA System    & 0.0811                  & 0.1454          & 0.0916             & 0.2449       & 0.0005        \\
MindLab Red Lions++  & 0.0830                  & 0.1469          & 0.0932             & 0.2394       & 0.0005        \\
AUEB-System5         & 0.0943                  & 0.1191          & 0.0892             & 0.2217       & 0.0011        \\
MindLab QA Reloaded  & 0.0605                  & 0.1103          & 0.0691             & 0.2106       & 0.0002        \\
Deep ML methods for  & 0.0815                  & 0.0931          & 0.0811             & 0.2051       & 0.0001        \\
bioinfo-0            & 0.1138                  & 0.1617          & 0.1175             & 0.1884       & 0.0009        \\
MindLab QA System ++ & 0.0639                  & 0.0990          & 0.0690             & 0.1874       & 0.0001        \\
AUEB-System3         & 0.0966                  & 0.1285          & 0.0935             & 0.1556       & 0.0011        \\
bio-answerfinder     & 0.0910                  & 0.1617          & 0.1004             & 0.1418       & 0.0008        \\
AUEB-System4         & 0.0080                  & 0.0082          & 0.0077             & 0.0328       & 0.0000       \\
        \hline
        \end{tabular}
        \caption{Results for snippet retrieval in batch 2 of phase A of Task 8b.
        }\label{tab:bA_res_sni}
\end{table*}

\textbf{Phase B}: 
In the second phase of task 8b, the participating systems were expected to provide both exact and ideal answers. 
Regarding the ideal answers, the systems will be ranked according to manual scores assigned to them by the BioASQ experts during the assessment of systems responses~\cite{balikas13}. 
For the exact answers, which are required for all questions except the summary ones, the measure considered for ranking the participating systems depends on the question type. 
For the yes/no questions, the systems were ranked according to the macro-averaged F1-measure on prediction of no and yes answer. 
For factoid questions, the ranking was based on mean reciprocal rank (MRR) and for list questions on mean F1-measure.
Some indicative results for exact answers for the third batch of Task 8b are presented in Table~\ref{tab:bB_res}. The full results of phase B of Task 8b are available online\footnote{\footnotesize \url{http://participants-area.bioasq.org/results/8b/phaseB/}}. These results are preliminary, as the final results for Task 8b will be available after the manual assessment of the system responses by the BioASQ team of biomedical experts.

\begin{table*}[!htbp]
\centering
\begin{tabular}
{M{0.205\linewidth}M{0.0852\linewidth}M{0.0852\linewidth}M{0.105\linewidth}M{0.11\linewidth}M{0.0852\linewidth}M{0.0852\linewidth}M{0.0852\linewidth}M{0.0852\linewidth}}
\hline

\textbf{System} & \multicolumn{2}{c}{\textbf{Yes/No}} & \multicolumn{3}{c}{\textbf{Factoid}} & \multicolumn{2}{c}{\textbf{List}} \\ 
\hline
& Acc. & F1 & Str. Acc. & Len. Acc. & MRR & Prec. & Rec. & F1 \\ \cline{2-9}    
Umass\_czi\_5         & \textbf{0.9032}  & 0.8995   & 0.2500   & 0.4286   & 0.3030   & \textbf{0.7361}   & 0.4833   & \textbf{0.5229}  \\
Umass\_czi\_1         & 0.8065  & 0.8046   & 0.2500   & 0.3571   & 0.2869   & 0.6806   & 0.4444   & 0.4683  \\
Umass\_czi\_2         & 0.8387  & 0.8324   & 0.2500   & 0.3571   & 0.2869   & 0.6806   & 0.4444   & 0.4683  \\
pa-base               & \textbf{0.9032}  & 0.8995   & 0.2500   & 0.4643   & 0.3137   & 0.5278   & 0.4778   & 0.4585  \\
pa                    & \textbf{0.9032}  & 0.8995   & 0.2500   & 0.4643   & 0.3137   & 0.5278   & 0.4778   & 0.4585  \\
Umass\_czi\_4         & \textbf{0.9032}  & 0.9016   & \textbf{0.3214}   & 0.4643   & 0.3810   & 0.6111   & 0.4361   & 0.4522  \\
KU-DMIS-1      & \textbf{0.9032} &\textbf{0.9028}   & \textbf{0.3214}   & 0.4286   & 0.3601   & 0.6583   & 0.4444   & 0.4520  \\
KU-DMIS-4      & 0.8387  & 0.8360   & 0.2857   & 0.4286   & 0.3357   & 0.6167   & 0.4444   & 0.4490  \\
KU-DMIS-5      & \textbf{0.9032}  & \textbf{0.9028}   & \textbf{0.3214}   & 0.4643   & 0.3565   & 0.6167   & 0.4444   & 0.4490  \\
KU-DMIS-2      & 0.8710  & 0.8697   & \textbf{0.3214}   & 0.4286   & 0.3446   & 0.6028   & 0.4444   & 0.4467  \\
KU-DMIS-3      & 0.8387  & 0.8360   & 0.2500   & 0.4643   & 0.3357   & 0.6111   & 0.4444   & 0.4431  \\
UoT\_allquestions     & 0.5806  & 0.3673   & \textbf{0.3214}   & 0.3929   & 0.3423   & 0.5972   & 0.4111   & 0.4290  \\
UoT\_baseline         & 0.5806  & 0.3673   & \textbf{0.3214}   & 0.3929   & 0.3512   & 0.4861   & 0.4056   & 0.4214  \\
Best factoid          & 0.5806  & 0.4732   & 0.2857   & 0.3929   & 0.3333   & 0.5208   & 0.4056   & 0.4107  \\
bio-answerfinder      & 0.8710  & 0.8640   & \textbf{0.3214}   & 0.4286   & 0.3494   & 0.3884   & \textbf{0.5083}   & 0.4078  \\
FudanLabZhu2          & 0.7419  & 0.6869   & \textbf{0.3214}   & 0.5357   &\textbf{0.3970}   & 0.5694   & 0.3583   & 0.3988  \\
FudanLabZhu3          & 0.7419  & 0.6869   &\textbf{0.3214}   & 0.4643   & 0.3655   & 0.5583   & 0.3472   & 0.3777  \\
FudanLabZhu4          & 0.7419  & 0.6869   & 0.2857   & \textbf{0.5714}   & 0.3821   & 0.5583   & 0.3472   & 0.3777  \\
FudanLabZhu5          & 0.7419  & 0.6869   & \textbf{0.3214}   & 0.4286   & 0.3690   & 0.5583   & 0.3472   & 0.3777  \\
UoT\_multitask\_l. & 0.5161  & 0.3404   & \textbf{0.3214}   & 0.4286   & 0.3643   & 0.5139   & 0.3556   & 0.3721  \\
BioASQ\_Baseline      & 0.5161  & 0.5079   & 0.0714   & 0.2143   & 0.1220   & 0.2052   & 0.4833   & 0.2562  \\

\hline
\end{tabular}
\caption{Results for batch 3 for exact answers in phase B of Task 8b.
Only the performance of the top-20 systems and the BioASQ Baseline are presented.
\label{tab:bB_res}}
\end{table*}

\begin{figure*}[!htbp]
\centerline{\includegraphics[width=1\textwidth]{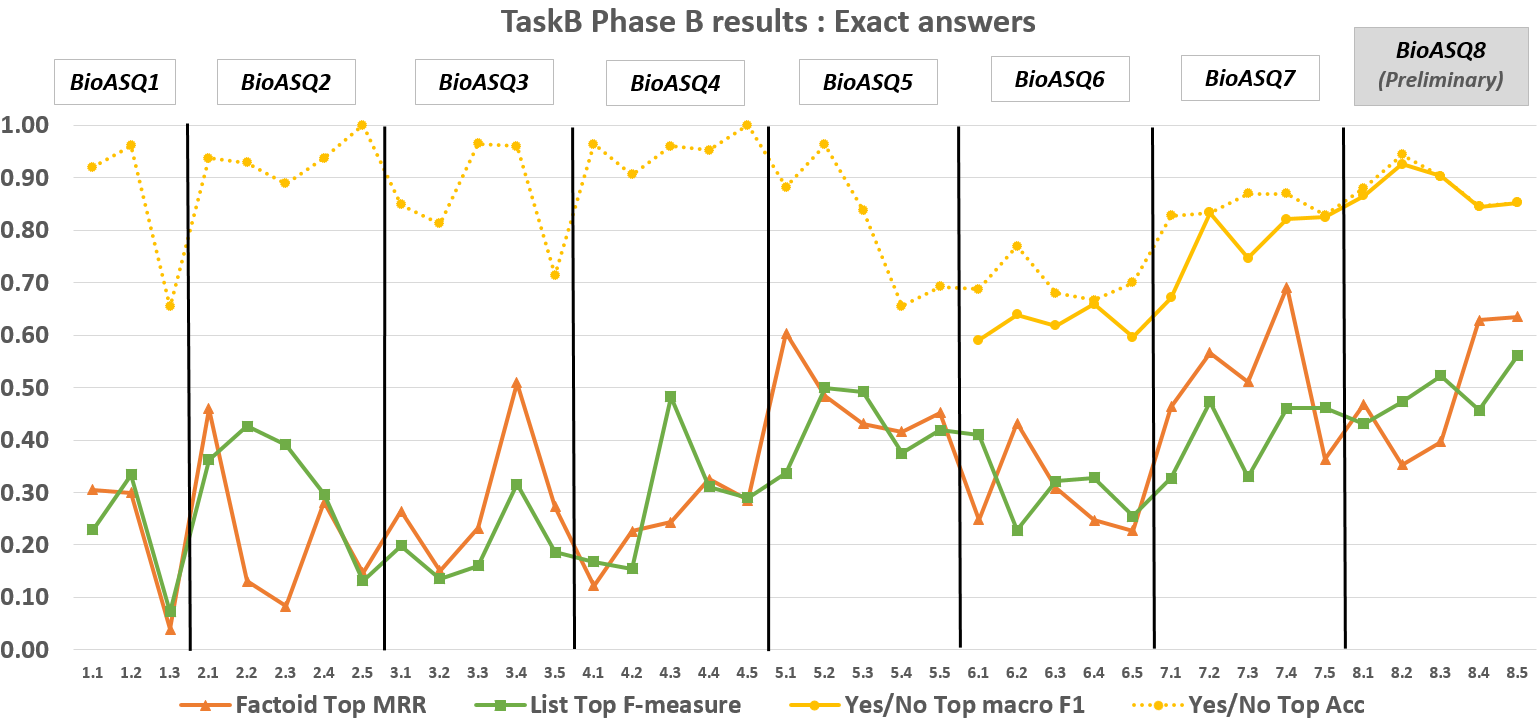}}
\caption{
The official evaluation scores of the best performing systems in Task B, Phase B, exact answer generation, across the eight years of the BioASQ challenge.
Since BioASQ6 the official measure for Yes/No questions is the macro-averaged F1 score (macro F1), but accuracy (Acc) is also presented as the former official measure. The results for BioASQ8 are preliminary, as the final results for Task 8b will be available after the manual assessment of the system responses. }\label{fig:Exact}
\end{figure*}

Figure {\ref{fig:Exact}} presents the performance of the top systems for each question type in exact answers during the eight years of the BioASQ challenge. 
The diagram reveals that this year the performance of systems in the yes/no questions keeps improving. For instance, in batch 3 presented in Table \ref{tab:bB_res}, various systems manage to outperform by far the strong baseline, which is based on a version of the OAQA system that achieved top performance in previous years.
Improvements are also observed in the preliminary results for list questions, whereas the top system performance in factoid questions is fluctuating in the same range as done last year. 
In general, Figure {\ref{fig:Exact}} suggests that for the latter types of question there is still more room for improvement.  

\subsection{Task MESINESP8}
The task proved to be a challenging one, but overall we believe the results were pretty good. Compared to the setting for English, the overall dataset was significantly smaller, and also the track evaluation contained not only medical literature, but also clinical trial summaries and healthcare project summaries. Moreover, in case of the provided training data, two different indexing approaches were used by the literature databases: IBECS has a more centralized manual indexing contracting system, while in case of LILACS a number of records were indexed in a sort of distributed community human indexer effort. The training set contained 23,423 unique codes, while the 911 articles in the evaluation set contained almost 4,000 correct DeCS codes. The best predictions, by Fudan University, scored a MIF (micro F-measure) of 0.4254 MiF using their AttentionXML with multilingual-BERT system, compared to the baseline score of  0.2695. Table \ref{tab:mesinesp_res} shows the results of the runs for this task. As a matter of fact, the five best scores were from them.
\begin{table}[!htb]
\centering
\begin{tabular}{M{0.31\linewidth}M{0.105\linewidth}M{0.105\linewidth}M{0.105\linewidth}M{0.105\linewidth}M{0.105\linewidth}M{0.105\linewidth}}
\hline
\textbf{System}     & MiF       		& MiP    & MiR & EBF       		  & MaF        & Acc.   \\
\hline

Model 4             & \textbf{0.4254}  & 0.4374       & \textbf{0.4140}     & \textbf{0.4240}    & \textbf{0.3194}  	   & \textbf{0.2786}        \\
Model 3             & 0.4227           & 0.4523       & 0.3966      		 & 0.4217             & 0.3122      			& 0.2768        \\
Model 1             & 0.4167           & 0.4466       & 0.3906      		 & 0.4160             & 0.3024      			& 0.2715        \\
Model 2             & 0.4165           & 0.4286       & 0.4051      		 & 0.4150             & 0.3082      			& 0.2707        \\
Model 5             & 0.4130           & 0.4416       & 0.3879      		 & 0.4122             & 0.3039      			& 0.2690        \\
PriberamTEnsemble   & 0.4093           & 0.5336       & 0.3320      		 & 0.4031             & 0.2115      			& 0.2642        \\
PriberamSVM         & 0.3976           & 0.4183       & 0.3789      		 & 0.3871             & 0.2543      			& 0.2501        \\
iria-mix            & 0.3892           & 0.5353       & 0.3057      		 & 0.3906             & 0.2318      			& 0.2530        \\
PriberamBert        & 0.3740           & 0.4293       & 0.3314      		 & 0.3678             & 0.2009      			& 0.2361        \\
iria-1              & 0.3630           & 0.5024       & 0.2842      		 & 0.3643             & 0.1957      			& 0.2326        \\
iria-3              & 0.3460           & 0.5375       & 0.2551      		 & 0.3467             & 0.1690      			& 0.2193        \\
iria-2              & 0.3423           & 0.4590       & 0.2729      		 & 0.3408             & 0.1719      			& 0.2145        \\
PriberamSearch      & 0.3395           & 0.4571       & 0.2700      		 & 0.3393             & 0.1776      			& 0.2146        \\
iria-4              & 0.2743           & 0.3068       & 0.2481      		 & 0.2760             & 0.2619      			& 0.1662        \\
BioASQ\_Baseline    & 0.2695           & 0.2337       & 0.3182      		 & 0.2754             & 0.2816      			& 0.1659        \\
graph matching      & 0.2664           & 0.3501       & 0.2150      		 & 0.2642             & 0.1422      			& 0.1594        \\
exact matching      & 0.2589           & 0.2915       & 0.2328      		 & 0.2561             & 0.0575      			& 0.1533        \\
LasigeBioTM TXMC F1 & 0.2507           & 0.3559       & 0.1936      		 & 0.2380             & 0.0858      			& 0.1440        \\
Anuj\_Ensemble      & 0.2163           & 0.2291       & 0.2049      		 & 0.2155             & 0.1746      			& 0.1270        \\
Anuj\_NLP           & 0.2054           & 0.2196       & 0.1930      		 & 0.2044             & 0.1744      			& 0.1198        \\
NLPUnique           & 0.2054           & 0.2196       & 0.1930      		 & 0.2044             & 0.1744      			& 0.1198        \\
X-BERT BioASQ F1    & 0.1430           & 0.4577       & 0.0847      		 & 0.1397             & 0.0220      			& 0.0787        \\
LasigeBioTM TXMC P  & 0.1271           & 0.6864       & 0.0701      		 & 0.1261             & 0.0104      			& 0.0708        \\
Anuj\_ml            & 0.1149           & \textbf{0.7557}       & 0.0621     & 0.1164             & 0.0006      		   	& 0.0636        \\
X-BERT BioASQ       & 0.0909           & 0.5449       & 0.0496       		 & 0.0916             & 0.0045      			& 0.0503       \\ \hline
\end{tabular}

\caption{ Final scores for MESINESP task submissions, including the official MiF metric in addition to other complementary metrics.
\label{tab:mesinesp_res}}
\end{table}

Although MiF represent the official competition metric, other metrics are provided for completeness. It is noteworthy that another team (Anuj-ml, from India) that was not among the highest scoring on MiF, nevertheless scored considerably higher than other teams with Precision metrics such as EBP (Example Based Precision), MaP (Macro Precision) and  MiP (Micro Precision). Unfortunately, at this time we have not received details on their system implementation.
One problem with the medical semantic concept indexing in Spanish, at least for diagnosis or disease related terms, is the uneven distribution and and high variability.~\cite{almagro2020}. 

\section{Conclusions}
\label{sec:conclusion}
This paper provides an overview of the eighth BioASQ challenge. This year, the challenge consisted of three tasks: The two tasks on biomedical semantic indexing and question answering in English, already established through the previous seven years of the challenge, and the new MESINESP task on semantic indexing of medical content in Spanish, which ran for the first time. 
The addition of the new challenging task on medical semantic indexing in Spanish, revealed that in a context beyond the English language, there is even more room for improvement, highlighting the importance of the availability of adequate resources for the development and evaluation of systems to effectively help biomedical experts dealing with non-English resources.

The overall shift of participant systems towards deep neural approaches, already noticed in the previous years, is even more apparent this year. 
State-of-the-art methodologies have been successfully adapted to biomedical question answering and novel ideas have been investigated. 
In particular, most of the systems adopted on neural embedding approaches, notably based on BERT and BioBERT models, for all tasks of the challenge.
In the QA task in particular, different teams attempted transferring knowledge from general domain QA datasets, notably SQuAD, or from other NLP tasks such as NER and NLI, also experimenting with multi-task learning settings. 
In addition, recent advancements in NLP, such as  XLNet~\cite{DBLP:journals/corr/abs-1906-08237}, BART~\cite{lewis2019bart} and SpanBERT~\cite{joshi2020spanbert} have also been tested for the tasks of the challenge. 

Overall, as in previous versions of the challenge, the top preforming systems were able to advance over the state of the art, outperforming the strong baselines on the challenging shared tasks offered by the organizers.
Therefore, we consider that the challenge keeps meeting its goal to push the research frontier in biomedical semantic indexing and question answering. 
The future plans for the challenge include the extension of the benchmark data though a community-driven acquisition process.

\section{Acknowledgments}
Google was a proud sponsor of the BioASQ Challenge in 2019. The eighth edition of BioASQ is also sponsored by the Atypon Systems inc. 
BioASQ is grateful to NLM for providing the baselines for task 8a and to the CMU team for providing the baselines for task 8b.
The MESINESP task is sponsored by the Spanish Plan for advancement of Language Technologies (Plan TL) and the Secretaría de Estado para el Avance Digital (SEAD).
BioASQ is also grateful to LILACS, SCIELO and Biblioteca virtual en salud and Instituto de salud Carlos III for providing data for the BioASQ MESINESP task.
%
%
%
\bibliographystyle{splncs04}
\bibliography{BioASQ8.bib}

\end{document}